\documentclass[12pt]{article}

\usepackage{coling2018}

\usepackage{url}
\usepackage{hyperref}
\usepackage{booktabs}
\usepackage{float}
\usepackage{graphicx}
\usepackage[table]{xcolor}
\usepackage{colortbl}
\usepackage{caption}
\usepackage{amsmath}
\usepackage[stable]{footmisc}

\usepackage{fontspec}
\newfontfamily\syriacfont[
    Path = fonts/,
    UprightFont = *-Regular,
    BoldFont = *-Bold,
    Extension = .ttf,
    Script = Syriac
]{NotoSansSyriacEastern}

\definecolor{headergray}{RGB}{220,220,220}

\date{}

\begin{document}

\title{Machine Translation between English and Syriac (East Syriac Dialect) using Statistical Machine Learning}

\author{
	\begin{tabular}[t]{c}
		Hadiana Sliwa and Hossein Hassani\\
		\textnormal{University of Kurdistan Hewl\^er}\\
		\textnormal{Kurdistan Region - Iraq}\\
		{\tt \{hadiana.harun, hosseinh\}@ukh.edu.krd}
	\end{tabular}
}

\maketitle

\begin{abstract}
UNESCO considers the Assyrian (Syriac) language an endangered language. Although Assyrians speak the language worldwide, the speaking population is uncertain (ranging from 500,000 to 1,500,000). Syriac is also one of the least studied languages in Natural Language Processing (NLP). Despite advances in Machine Translation (MT) over the past decade, the lack of publicly available corpora and the orthographic complexity of the Syriac script, specifically the Madnkhaya script, have left this language entirely ignored in the computational linguistics literature. This study develops the first phrase-based Statistical MT (SMT) model for English-to-Assyrian MT using the Moses framework. We created a dataset of 38,847 sentence pairs from the complete English and Syriac Bible, merging a pre-existing New Testament dataset with an Old Testament built from scratch through PDF extraction, using custom segmentation scripts and manual alignment review by three bilingual annotators. The Syriac side of the corpus undergoes diacritic removal and Byte-Pair Encoding tokenization to reduce orthographic sparsity before training. We trained and evaluated six models using different configurations and splitting-scheme ratios, language model order, distortion limits, and the inclusion of an Operation Sequence Model. The best-performing configuration achieves a word-level BLEU score of 23.54. Human evaluation by 11 native Assyrian speakers resulted in mean adequacy and fluency scores of 3.42 and 3.34 out of 5, respectively. These results are consistent with comparable low-resource SMT models trained on Biblical corpora for morphologically rich Semitic languages. The corpora, scripts, and trained model are publicly available, providing the research community with the first systematically curated English--Syriac dataset and a reproducible baseline for future MT and broader NLP work on this endangered language.
\end{abstract}

\section{Introduction}
The digital era has seen a revolutionary shift in how we communicate across linguistic barriers, primarily driven by the advancement of Machine Translation (MT). However, this progress has not been equal for all languages. Although high-resource languages like English and French have close to human-quality translation \cite{hassan2018achievinghumanparityautomatic}, low-resource languages such as Kurdish \cite{AhmadiHassaniJaff2022}, Assyrian (Syriac) \cite{Transformer_based_Parser_for_Syriac}, and many other languages are left behind. Some of these languages have little to no presence on the internet, with rich texts and scriptures still on physical documents stored in places such as churches, mosques, and other historical sites. Although research on Assyrian handwritten character recognition has reached encouraging results, achieving a maximum accuracy of 95.70\% with the MobileNet-V2 model \cite{Armya_Abdulrazzaq_2024,majeed2024ancient}, such models still need further development before full handwritten documents can be reliably digitized.

A further limitation specific to the Assyrian language is ``font-based transliteration'', where digitized documents are written with software such as MS Word using Arabic characters, with a Syriac font applied on top so that the Arabic Unicode characters visually display Assyrian (Syriac) letters \cite{Kiraz_2011}. The underlying encoding remains Arabic; only the glyphs are swapped. Currently, no software or script reliably converts this Arabic-encoded text into actual Syriac Unicode. This practice originated before a Unicode range existed for Syriac, and adoption of the proper encoding has remained limited due to unfamiliar keyboard layouts and the lack of default system support for Syriac in operating systems and software.

The main challenge addressed in this work is the extreme scarcity of linguistic data for the Assyrian language, which is classified by UNESCO as an endangered language. Historically, Syriac speakers mainly resided in Middle Eastern countries, including Iran, Iraq, Syria, and Turkey. However, instability in the region has led to large-scale migration to countries such as the United States, Australia, and Canada, where the new generations are increasingly less likely to learn to speak, read, or write the language. Most of the available online material is in non-Unicode formats or scanned documents, and the lack of reliable OCR and text-processing tools for Syriac significantly limits large-scale digitization and preservation.

This research investigates whether a pipeline of devocalization and orthographic normalization can bridge the data gap in low-resource settings, using the traditionally data-efficient Moses Statistical Machine Translation (SMT) framework to establish the first reproducible baseline for the English--Syriac pair.

The rest of this paper is organized as follows: Section~\ref{sec:syriac} gives a brief overview of the Syriac language and script, Section~\ref{sec:related} reviews related work, Section~\ref{sec:method} describes the method, Section~\ref{sec:results} presents the results and discussion, and Section~\ref{sec:conclusion} concludes.

\section{The Syriac Language}
\label{sec:syriac}

Syriac is a form of Aramaic belonging to the Northwest Semitic group of languages, a group that also includes Hebrew, Ugaritic, and Phoenician; within the family tree, Aramaic and Arabic are grouped together under ``Central Semitic'', distinct from East Semitic (Akkadian) and South Semitic (Ethiopic) \cite{Brock2017}. Its historical and religious significance lies in its role as a foundational language of early Syriac Christianity and as a ``bridge culture'' through which Greek philosophy, medicine, and science were transmitted to the Islamic world and eventually to Western Europe \cite{Brock2017,Muraoka2005}.

Syriac has two main dialects, West Syriac and East Syriac \cite{Kiraz2013}. These are not separate languages, but dialects that are mostly identical in grammar and lexicon, with differences that are largely phonological and orthographic. This work focuses on East Syriac, used by the Church of the East and Chaldean communities, written in the Madnkhaya script, which emerged around the 6th century.

The Syriac alphabet is essentially consonantal, consisting of twenty-two letters written and read from right to left in a cursive, joined-up style \cite{Muraoka2005}. Three letters (Alap {\syriacfont ܐ}, Waw {\syriacfont ܘ}, and Yodh {\syriacfont ܝ}) are bivalent: they can function as consonants, act as vowel letters, or, in the case of Alap, be entirely silent. Six plosive consonants possess a twofold (hard/soft) pronunciation governed by the \textit{Rukakha} and \textit{Qushshaya} dots, and eight letters join only to the right, never connecting to the letter that follows them. Letter forms also change depending on their position within a word, and several letter pairs are visually similar and easily confused (e.g., Dalath {\syriacfont ܕ} vs.\ Resh {\syriacfont ܪ}).

East Syriac orthography uses seven vowel markers (West Syriac uses five), realised as combinations of dots placed above or below consonants, as summarised in Table~\ref{tab:syriac_dic_func}. In addition to vowels, the diacritic system includes \textit{Sey\={a}m\={e}} (a double dot marking plural nouns, e.g.\ {\syriacfont ܡ̈ܠܟܐ} /malk\={e}/ `kings' vs.\ {\syriacfont ܡܠܟܐ} /malk\={a}/ `king') and \textit{M\d{t}alq\={a}n\={a}} (a diagonal line marking a silent letter) \cite{Muraoka2005}. In everyday writing, native readers and writers routinely omit vowel diacritics, relying on consonantal context and morphological knowledge, a foundational characteristic of Semitic writing systems shared with Arabic and Hebrew. As Section~\ref{sec:preprocessing} explains, this convention directly motivates the devocalization step in our preprocessing pipeline, because the same underlying word may appear fully vocalized, partially vocalized, or unvocalized in the source data.

\begin{table}[h]
\centering
\caption{East Syriac vowel diacritics and their functions}
\small
\renewcommand{\arraystretch}{1.3}
\begin{tabular}{|l|c|l|l|}
\hline
\rowcolor{headergray}
\textbf{Pronunciation} & \textbf{Mark} & \textbf{Name} & \textbf{Function / Value} \\
\hline
/Ba/ & {\syriacfont ܒܲ} & Ptakha & short /a/ vowel \\
/B$\bar{a}$/ & {\syriacfont ܒܵ} & Zqapha & long /\={a}/ vowel \\
/B\textit{i}/ & {\syriacfont ܒܸ} & Zlameh Kirye & short /i/ sound \\
/B$\bar{e}$/ & {\syriacfont ܒܹ} & Zlameh Yareekhe & /$\epsilon$/ sound \\
/Bo/ & {\syriacfont ܒܘܿ} & Rwakha & O sound \\
/Boo/ & {\syriacfont ܒܘܼ} & Rwasa & oo or u sound \\
/Bee/ & {\syriacfont ܒܝܼ} & Khwasa & ee sound \\
\hline
\end{tabular}
\label{tab:syriac_dic_func}
\end{table}

Morphologically, Syriac follows the Semitic root-and-pattern system: a single root can generate a large number of surface word forms through prefixation, suffixation, and cliticization, and nouns, adjectives, and verbs must agree in gender and number with their nucleus noun \cite{Muraoka2005}. Syntactically, Classical Syriac frequently uses Verb--Subject constructions, diverging from the Subject--Verb--Object order of English, a structural divergence with direct consequences for word alignment, discussed in Section~\ref{sec:smt_config}.

\section{Related Work}
\label{sec:related}

This section addresses research from two intersecting areas relevant to the current study: MT for low-resource languages and computational work on Syriac and its linguistically adjacent languages. We do not focus on high-resource languages, which mostly now use Neural Machine Translation because the research community is well aware of their status and they rely on very large-scale corpora or pre-trained models, which are unavailable for Syriac.  

\subsection{Machine Translation in Low-Resource Languages}
MT has undergone a revolutionary transformation over the previous decades, evolving from rule-based systems to SMT, and then to the highly fluent Neural Machine Translation (NMT) models used today. Many approaches have been taken to improve MT in low-resource settings, such as active learning, data augmentation, and transfer learning, but the scarcity of data remains the critical constraint \cite{article_tafa_2025}.

\newcite{kumar-etal-2021-machine} focused on adapting MT systems to low-resource language varieties and typologically related languages, proposing a transfer-learning framework named LangVarMT that adapts a model trained on a ``standard'' variety of a language to a related low-resource variety through embedding mapping, vocabulary recycling, back-translation, and a final training pivot. The framework outperformed all competitive baselines, including Ukrainian, Belarusian, Nynorsk, and several Arabic dialects; with only 10,000 monolingual sentences, it surpassed the strongest baselines by more than four BLEU points for Ukrainian. However, its success depends heavily on the relatedness of the standard and target varieties, and it cannot function for a language --in our case, the Syriac script-- that has no suitable pre-trained ``bridge'' model.

\newcite{velayuthan-etal-2024-back} showed the importance of data quality over quantity for low-resource pairs. Although the final model was neural, statistical methods drove their bilingual filtering pipeline, merging rule-based cleaning with Jensen-Shannon Divergence to select high-quality samples. The model reached maximum performance at 100K sentences (50.04 chrF, 18.17 BLEU) before performance declined as a result of adding more low-quality data.

When datasets are limited, SMT is often preferred over NMT \cite{koehn-knowles-2017-six}: it does not require large amounts of data and takes less time to train. \newcite{kakum_22_nyishi} implemented a phrase-based SMT system for Nyishi, a low-resource and endangered Indian language, collecting 30,000 pairs of sentences of English--Nyishi, cross-checked by native elders and language experts, and using the Moses toolkit, MGIZA, and IRSTLM toolkit. Tuning on 967 sentence pairs increased BLEU from 0.1419 to 0.1849 for Nyishi--English and from 0.0802 to 0.1411 for English--Nyishi.

\subsection{Machine Translation in Less-Studied Languages Close to Syriac}
At the time of writing, computational research dedicated specifically to Syriac in the MT domain remains remarkably scarce. An exception is \newcite{naaijer-etal-2023-transformer}, in which the authors developed a transformer-based sequence-to-sequence model to parse the morphology of Ancient Syriac, training on 5,596 Syriac verses from the ETCBC database augmented with 22,946 verses of the Hebrew Masoretic Text. The model trained on Syriac data alone reached 89.3\% accuracy, and the best Hebrew-augmented model reached 90.8\%, a modest gain considering that the Hebrew data is almost four times larger than the Syriac dataset and substantially increases training cost.

Closer to our work, \newcite{Chaya_2024_AFCM} built a publicly available Biblical parallel Aramaic--Hebrew corpus covering the entire Hebrew Bible and compared phrase-based SMT against RNN and Transformer NMT architectures. The results were impressive: SMT outperformed NMT in this low-resource context, achieving a BLEU score of 44.3 compared to NMT's 35.8, with the lack of training data cited as the main obstacle for the neural models. Similarly, \newcite{Guellil_2007} developed a two-step framework that first transliterates Arabizi into Arabic script and then translates Algerian dialect into Modern Standard Arabic. While the neural approach won on transliteration, SMT outperformed NMT on the actual translation task (6.01 BLEU vs.\ scores often at 0.0), on a corpus of only 6,412 parallel sentences.

To summarize, across the reviewed literature, a consistent picture emerges: in low-resource settings, SMT is the preferred approach due to the lower data requirements, faster training, and robustness to sparse and noisy datasets \cite{koehn-knowles-2017-six,Chaya_2024_AFCM,Guellil_2007}. Since large-scale datasets for the Assyrian language are not available, this study follows the SMT approach. 

\section{Method}
\label{sec:method}

\begin{figure}[htbp]
    \centering
    \includegraphics[height=0.90\textheight]{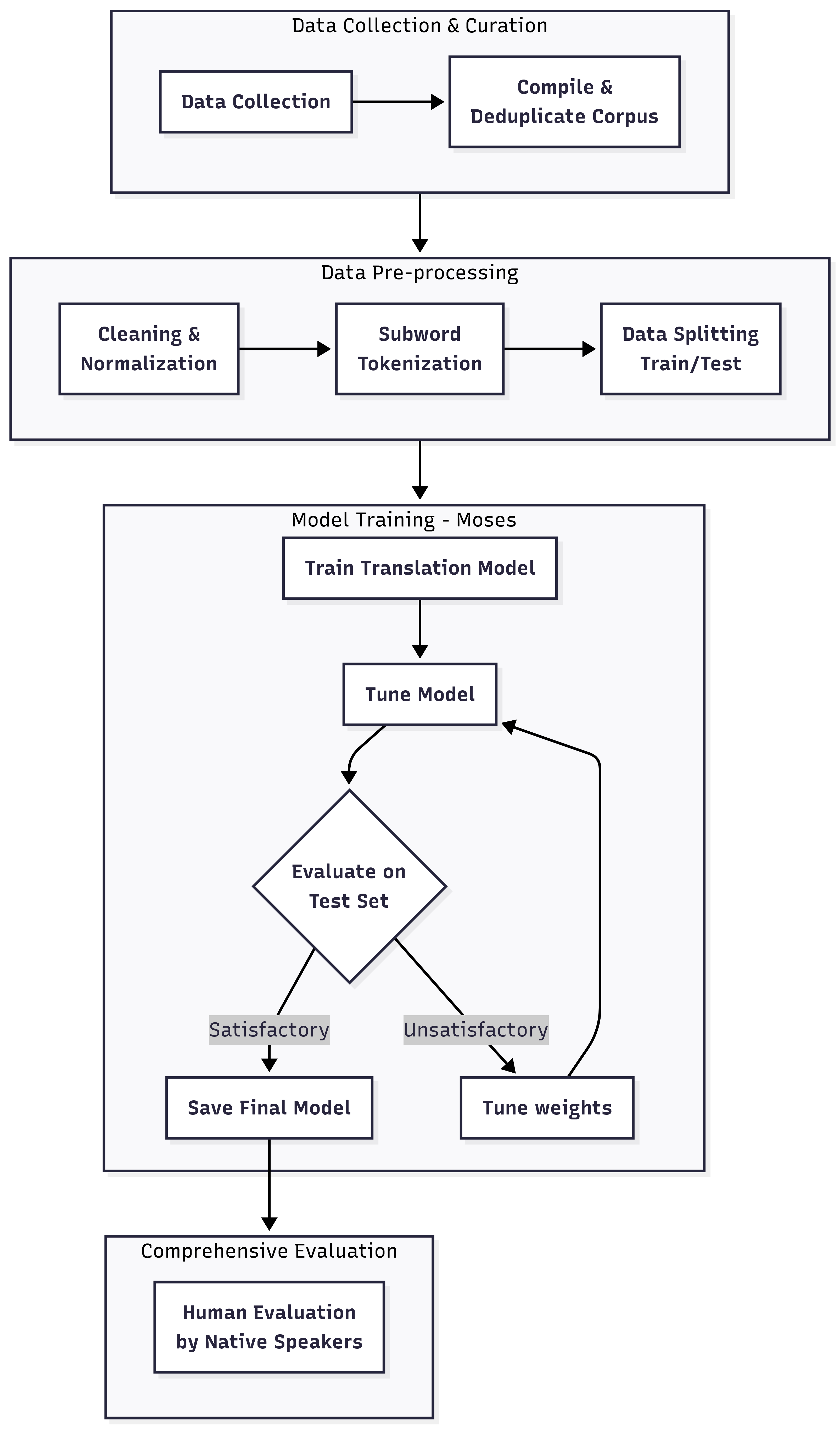}
    \caption{Overview of the research method}
    \label{fig:model_training}
\end{figure}

This research uses the Moses phrase-based SMT framework. 
Figure~\ref{fig:model_training} shows the overall flow of the method, with each block explained in the following subsections.

\subsection{Data Collection and Curation}
To address the lack of credible sources for Syriac translation, we collect data from sources such as Biblical and educational references. This stage includes data curation, corpus compilation, and deduplication. 

\subsection{Data Preprocessing}
\label{sec:preprocessing}
 The stage consists of cleaning and normalization, devocalization, Byte-Pair Encoding tokenization of subwords, and data splitting.

\textbf{Cleaning and normalization --} Noise removal is applied to both sides of the corpus: a custom cleaning script (e.g., using Python) strips stray punctuation, symbols, inconsistent whitespace, and formatting characters introduced during extraction, and removes empty, duplicate, or one-sided sentence pairs. The most linguistically significant step at this stage is Syriac diacritic removal. As discussed in Section~\ref{sec:syriac}, native readers routinely omit vowel diacritics in everyday text, although the same underlying word may appear in multiple surface forms depending on vocalization. This sparsity is particularly problematic in low-resource settings, where each unique surface form competes for a limited training signal. We therefore apply a devocalization script that removes all diacritical marks from the Syriac data using their Unicode code points, creating a consistent consonantal representation, consistent with established practice in Semitic MT preprocessing, where vowel reduction is a standard step for Arabic and Hebrew \cite{Habash2006Preprocessing}.

\textbf{Subword tokenization --} We apply Byte-Pair Encoding (BPE) on both sides of the corpus. BPE iteratively merges the most frequent character-sequence pairs, producing a subword vocabulary that sits between character-level and word-level representations \cite{sennrich2016subword}. This suits the morphological characteristics of Syriac, which generates a large number of surface forms from a relatively small set of roots; BPE decomposes rare or unseen words into known subword units, substantially reducing the out-of-vocabulary problem.

\textbf{Data splitting --} After tokenization, the corpus is partitioned into training and test sets. To assess how translation performance scales with training data size, we train and evaluate under four split configurations: 95/5, 90/10, 80/20, and 70/30. Evaluating across multiple training sizes is an established methodology in low-resource MT research \cite{kakum_22_nyishi,velayuthan-etal-2024-back}. All splits are created by random sampling without replacement, so test sentences are never seen during training. Moses performs tuning internally via MERT on a held-out tuning portion, so no separate development set is required.

\subsection{Model Training}
\label{sec:smt_config}
The Moses pipeline integrates three interdependent components: a word alignment model, a phrase translation table, and a target-language model. The overall configuration is summarized in Table~\ref{tab:moses_config}.

\textbf{Word alignment with GIZA++ --} We use GIZA++, which implements the IBM alignment models 1--5 using the Expectation-Maximization algorithm \cite{AlOnaizan1999SMT}, run independently in both translation directions and symmetrized with the grow-diag-final-and heuristic, the Moses default shown to produce the best balance between alignment precision and recall \cite{koehn2003statistical}. Alignment is particularly challenging for the English--Syriac pair: English follows SVO order while Syriac frequently uses Verb--Subject constructions, and Syriac's prefixed prepositions and pronominal suffixes mean a single Syriac token may correspond to multiple English words. The devocalization applied earlier reduces surface variation and directly supports alignment quality.

\textbf{Phrase table construction --} Moses extracts all contiguous bilingual phrase pairs consistent with the word alignments up to a maximum phrase length, computing five feature scores per pair (forward and reverse translation probabilities, forward and reverse lexical weights, and a phrase penalty), combined in Moses's log-linear model whose weights are optimized during tuning.

\textbf{Language model --} We train a target-side KenLM language model on the Syriac side of the training corpus with modified Kneser-Ney smoothing, which handles the sparse count distributions characteristic of low-resource datasets effectively. A trigram order is selected as appropriate for the baseline corpus size; higher-order models require more data to estimate reliably.

\begin{table}[h]
\centering
\small
\renewcommand{\arraystretch}{1.3}
\begin{tabular}{|l|l|}
\hline
\rowcolor{headergray}
\textbf{Component} & \textbf{Setting} \\
\hline
Decoder                  & Moses phrase-based SMT \\
Alignment tool           & GIZA++ (IBM Models 1--5) \\
Symmetrization heuristic & grow-diag-final-and \\
Language model toolkit   & KenLM \\
Language model order     & 3-gram (baseline) / 5-gram (enhanced) \\
Smoothing                & Modified Kneser-Ney \\
Tuning algorithm         & MERT \\
Evaluation metric        & BLEU \\
\hline
\end{tabular}
\caption{Moses SMT system configuration}
\label{tab:moses_config}
\end{table}

Once the phrase table and language model are created, the log-linear model is tuned with MERT, which decodes a held-out tuning set repeatedly, compares the hypotheses against reference translations, and iteratively updates the feature weights to maximize BLEU. The full pipeline, alignment, phrase extraction, language model integration, and tuning, is coordinated through Moses's \texttt{train-model.perl} script and executed locally; no cloud or GPU resources are required, which is one of the practical advantages of phrase-based SMT for low-resource settings \cite{koehn-knowles-2017-six}. Training runs are performed independently for each of the four data splits, producing separate trained models for comparative evaluation.

\subsection{Evaluation}
We evaluate the trained models using two complementary approaches. The main automatic metric is BLEU \cite{papineni-etal-2002-bleu}, which measures n-gram overlap between hypothesis and reference translations with a brevity penalty. Despite its well-documented limitations, insensitivity to synonymy and weaker correlation with human judgment in morphologically rich languages, BLEU remains the standard benchmark in SMT research and enables direct comparison against related low-resource Semitic MT work. BLEU is computed with the Moses scoring scripts across all four split configurations.

Automatic metrics alone are insufficient for a morphologically complex, low-resource language such as Syriac, where reference translations may themselves exhibit variation \cite{human_eva_2021}. We therefore complement BLEU with a structured human evaluation by a panel of bilingual native Assyrian speakers. Evaluators receive a standardized assessment form containing English source sentences drawn from web articles not seen during training, alongside their model-generated Syriac translations, and rate each translation on a 5-point scale along two dimensions: \textit{Adequacy} (the degree to which the translation preserves the meaning of the source) and \textit{Fluency} (the degree to which the translation reads as natural, grammatically well-formed Syriac). This two-dimensional framework is standard in MT evaluation and separates meaning preservation from output naturalness.

\section{Results and Discussion}
\label{sec:results}
\subsection{Data Collection}
We collected data from biblical sources. The selection of Biblical text as the primary data source is motivated by practical and linguistic factors. Practically, the Bible is one of the very few texts for which a complete verse-aligned parallel exists between English and Syriac, owing to the Peshitta's central role in the Assyrian Christian tradition. Linguistically, Biblical Syriac uses the morphological and orthographic features of the Madnkhaya script that this research targets. The domain limitation is acknowledged: the resulting model is suited to the vocabulary of the Biblical text and is not intended as a general-purpose translation system, consistent with comparable low-resource Semitic MT research where domain-specific corpora are used when general corpora are unavailable \cite{Chaya_2024_AFCM}.

For the New Testament portion ($\sim$7,900 verses), we curate pre-existing verse-aligned English--Syriac text identified in academic and religious digital archives, passed through the same cleaning and normalization pipeline as the rest of the corpus. The Old Testament ($\sim$23,000 verses) presents a considerably greater challenge: no pre-existing digital parallel corpus is available in a clean, verse-aligned format, so we build this portion from scratch in four stages. First, raw text is extracted from the source documents using Python-based PDF extraction scripts. Second, a custom segmentation script uses the chapter-and-verse numbering system present in both Biblical sources (e.g., 1:1, 1:2) as anchor points to split the continuous text into discrete verse-level segments, an established practice in Biblical corpus construction, where verse numbers serve as natural and reliable alignment anchors across language versions. Third, the segmented verse pairs are combined into a single parallel file, one matched English--Syriac pair per line. Fourth, a manual review phase corrects the noise, misalignment, and formatting errors that automated PDF extraction inevitably introduces, particularly when processing right-to-left Syriac script.

We also collected data from the Ministry of Education (the Kurdistan Regional Government in the Kurdistan Region of Iraq). The collected documents are written in the same Madnkhaya script. However, we received them after the deadline for completing this research, so we could not incorporate them into our model. This data source includes non-liturgical, secular Syriac text, a domain almost absent from existing digital resources for this language. These materials are being processed through the same normalization pipeline and, unlike the Biblical corpus, they will not be released publicly: under the terms agreed with the Ministry, they are shared privately with individual researchers on written request, for academic research purposes only, and are randomised at sentence level so that they cannot be reconstructed as a teaching curriculum. Requests may be directed to the corresponding
author.

\subsection{Corpus creation}
We create a parallel English--Assyrian corpus of 38,847 aligned sentence pairs extracted from Biblical text.

Three individuals, the primary researcher and two collaborators with working knowledge of both English and Syriac, reviewed the aligned corpus. The Old Testament corpus is divided into sections, each inspected independently by a single reviewer, who corrects misalignments, removes pairs where extraction has failed, and flags verses where the Syriac text is incomplete. This division of labor is practical in low-resource NLP contexts where annotation resources are limited, and the resulting corpus represents a human-verified foundation for model training.

\begin{figure}[htbp]
    \centering
    \includegraphics[width=0.98\textwidth]{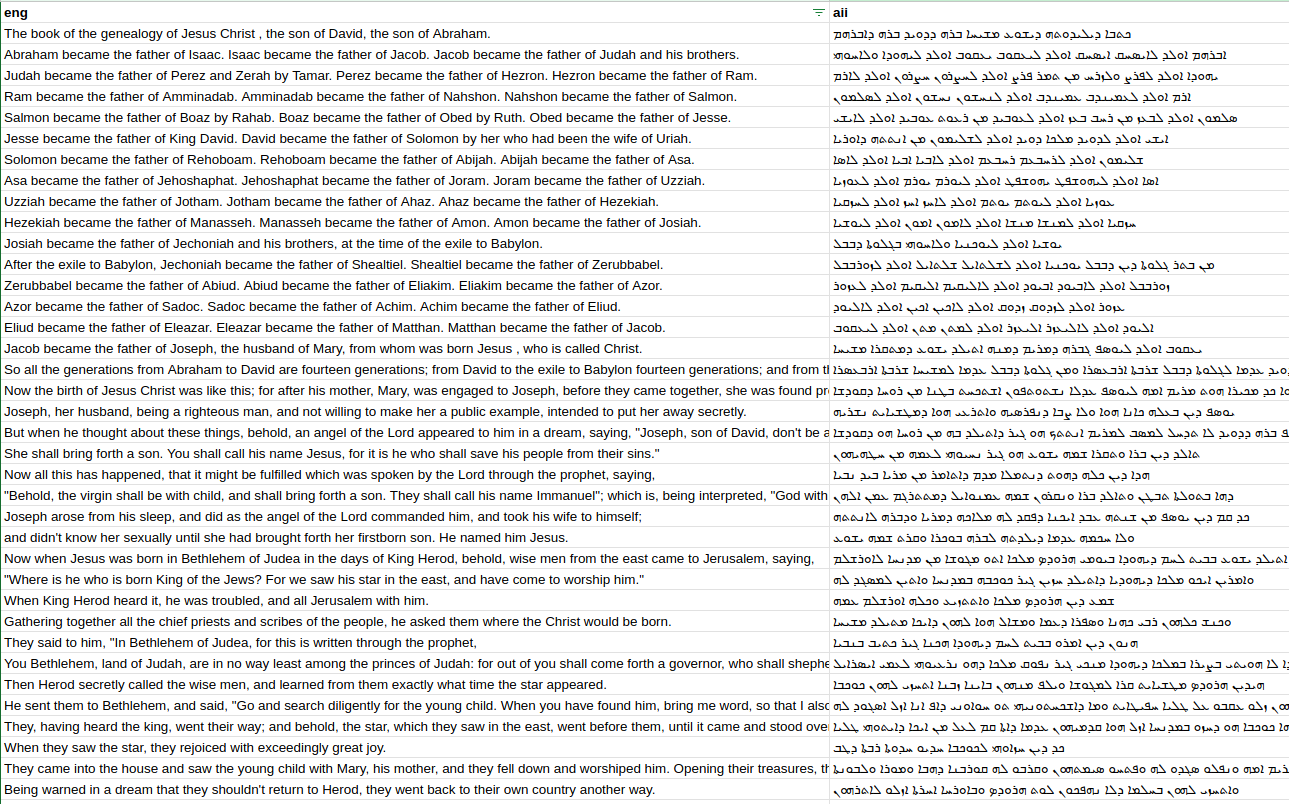}
    \caption{Sample from the curated dataset}
    \label{fig:sample_dataset_eng_aii}
\end{figure}

The parallel corpus contains 38,847 sentence pairs from the complete English and Syriac Bible; a sample is shown in Figure~\ref{fig:sample_dataset_eng_aii}. Following preprocessing with \texttt{clean-corpus-n.perl}, sentence pairs where either side exceeded 100 BPE tokens were removed. The corpus was shuffled with a fixed seed of 42 for reproducibility. A fixed tuning set of 1,000 sentences was held out for MERT across all models, and the remaining data was partitioned according to the four split ratios. Table~\ref{tab:corpus_stats} summarizes the corpus and split statistics, and Table~\ref{tab:model_configs} summarizes the six model configurations: Models 1--4 form the baseline set trained on identical configurations across four data splits, while Models 5 and 6 use the best-performing split (95/5) with architectural enhancements.

\subsection{Experimental Setup}

The experiments were performed according to Table~\ref{tab:corpus_stats}.

\begin{table}[h]
\centering
\small
\renewcommand{\arraystretch}{1.3}
\begin{tabular}{|l|c|c|c|c|}
\hline
\rowcolor{headergray}
\textbf{Model} & \textbf{Split} & \textbf{Train} & \textbf{Tune} & \textbf{Test} \\
\hline
Model 1 & 95/5  & 35,905 & 1,000 & 1,942  \\
Model 2 & 90/10 & 33,963 & 1,000 & 3,884  \\
Model 3 & 80/20 & 30,078 & 1,000 & 7,769  \\
Model 4 & 70/30 & 26,193 & 1,000 & 11,654 \\
Model 5 & 95/5  & 35,905 & 1,000 & 1,942  \\
Model 6 & 95/5  & 35,905 & 1,000 & 1,942  \\
\hline
\multicolumn{2}{|l|}{\textbf{Total sentence pairs}} & \multicolumn{3}{c|}{38,847} \\
\hline
\end{tabular}
\caption{Corpus and data split statistics}
\label{tab:corpus_stats}
\end{table}

\begin{table}[h]
\centering
\small
\renewcommand{\arraystretch}{1.3}
\begin{tabular}{|c|l|c|c|c|c|c|}
\hline
\rowcolor{headergray}
\textbf{Model} & \textbf{Description} & \textbf{LM} & \textbf{Distortion} & \textbf{BPE} & \textbf{OSM} & \textbf{Split} \\
\hline
1 & Baseline          & 3-gram & 6  & 10k & No  & 95/5  \\
2 & Baseline          & 3-gram & 6  & 10k & No  & 90/10 \\
3 & Baseline          & 3-gram & 6  & 10k & No  & 80/20 \\
4 & Baseline          & 3-gram & 6  & 10k & No  & 70/30 \\
5 & Enhanced (OSM)    & 5-gram & 6  & 10k & Yes & 95/5  \\
6 & Enhanced (dl/BPE) & 5-gram & 12 & 20k & No  & 95/5  \\
\hline
\end{tabular}
\caption{Summary of all model configurations}
\label{tab:model_configs}
\end{table}

\subsection{Automatic Evaluation Results}
\begin{figure}[htbp]
    \centering
    \includegraphics[width=0.98\textwidth]{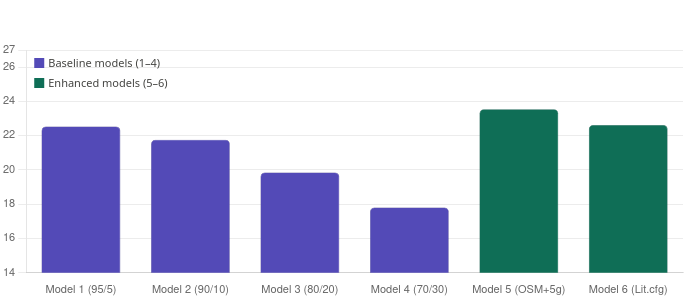}
    \caption{Word-level BLEU scores across all six model configurations}
    \label{fig:blue_scores}
\end{figure}

Table~\ref{tab:bleu_results} reports the word-level BLEU scores for all six models, visualized in Figure~\ref{fig:blue_scores}. All reported BLEU scores are computed at the word level using \texttt{multi-bleu.perl, a script in Moses} after de-BPE post-processing; BPE-level BLEU is shown for diagnostic purposes only, as BPE tokenization artificially inflates n-gram precision by shortening tokens \cite{sennrich2016subword}.

\begin{table}[h]
\centering
\small
\renewcommand{\arraystretch}{1.3}
\resizebox{\textwidth}{!}{%
\begin{tabular}{|c|l|l|c|c|c|}
\hline
\rowcolor{headergray}
\textbf{Model} & \textbf{Model/Split} & \textbf{Configuration} & \textbf{Test Size} & \textbf{Word BLEU} & \textbf{BPE BLEU*} \\
\hline
1 & Baseline 95/5  & 3-gram, dl=6, 10k BPE, MERT & 1,942  & 22.54 & 23.95 \\
2 & Baseline 90/10 & 3-gram, dl=6, 10k BPE, MERT & 3,884  & 21.76 & 22.91 \\
3 & Baseline 80/20 & 3-gram, dl=6, 10k BPE, MERT & 7,769  & 19.85 & 20.93 \\
4 & Baseline 70/30 & 3-gram, dl=6, 10k BPE, MERT & 11,654 & 17.81 & 18.84 \\
5 & Enhanced       & OSM, 5-gram, dl=6, 10k BPE, MERT & 1,942  & \textbf{23.54} & 24.60 \\
6 & Enhanced       & 5-gram, dl=12, 20k BPE, MERT & 1,942  & 22.63 & 22.64 \\
\hline
\multicolumn{6}{|l|}{\textit{* BPE BLEU shown for diagnostic comparison only; not used for evaluation}} \\
\hline
\end{tabular}
}
\caption{BLEU results for all six models}
\label{tab:bleu_results}
\end{table}

The baseline results confirm the expected monotonic relationship between training data volume and translation quality: BLEU decreases from 22.54 at the 95/5 split to 17.81 at the 70/30 split, a drop of 4.73 points as the training set shrinks from 35,905 to 26,193 sentence pairs. This pattern is consistent with findings across low-resource SMT research, where data volume is the primary ceiling on translation quality \cite{kakum_22_nyishi,velayuthan-etal-2024-back}.

Table~\ref{tab:bleu_breakdown} provides the n-gram precision breakdown for the best-performing model (Model 5). The brevity penalty of 1.000 confirms that Model 5 does not produce systematically shorter translations than the reference. The unigram precision of 48.7 indicates that approximately half of the words produced match the reference, a reasonable outcome for a morphologically rich target language where surface-form variation is high even for semantically equivalent outputs. The drop from unigram to 4-gram precision (48.7 $\rightarrow$ 12.7) reflects the difficulty of producing exact phrase sequences in Syriac, where inflectional variation means that even correct translations may not match the reference at the 3- and 4-gram level.

\begin{table}[h]
\centering
\small
\renewcommand{\arraystretch}{1.3}
\begin{tabular}{|c|c|c|c|c|c|c|}
\hline
\rowcolor{headergray}
\textbf{BLEU} & \textbf{1-gram} & \textbf{2-gram} & \textbf{3-gram} & \textbf{4-gram} & \textbf{BP} & \textbf{Ratio} \\
\hline
23.54 & 48.7 & 27.5 & 18.1 & 12.7 & 1.000 & 1.025 \\
\hline
\end{tabular}
\caption{N-gram BLEU breakdown for Model 5 (word-level)}
\label{tab:bleu_breakdown}
\end{table}

\textbf{Effect of model enhancements --} Model 5, which adds an Operation Sequence Model (OSM) and upgrades the LM from 3-gram to 5-gram, achieves the highest word-level BLEU of 23.54, an improvement of +1.00 over the identical-split baseline. The OSM provides an explicit reordering feature that captures longer-range word order changes through operation sequences \cite{durrani-etal-2011-joint}, particularly suitable for the English--Syriac pair given its SVO-to-VS structural divergence \cite{Carpuat2010Reordering}, while the 5-gram KenLM contributes more fluent target-side generation. Model 6, which increases the distortion limit from 6 to 12 and doubles BPE merge operations to 20,000, achieves 22.63 BLEU, a gain of only +0.09 over the baseline, suggesting that long-range reordering beyond the default limit is not the primary bottleneck for this corpus, and that the single-register nature of the Biblical training data imposes a ceiling that additional reordering capacity cannot overcome.

\subsection{Human Evaluation Results}
Twenty English sentences translated by Model 5, drawn from online articles covering Biblical topics not in the training data, were each rated by eleven native Assyrian speakers on adequacy and fluency (5-point scales), yielding 220 individual ratings per dimension. Table~\ref{tab:human_eval_overall} presents the aggregate statistics.

\begin{table}[h]
\centering
\small
\renewcommand{\arraystretch}{1.3}
\begin{tabular}{|l|c|}
\hline
\rowcolor{headergray}
\textbf{Metric} & \textbf{Value} \\
\hline
Total individual ratings        & 220 \\
Mean adequacy score             & 3.42 / 5 \\
Mean fluency score              & 3.34 / 5 \\
Mean overall score              & 3.38 / 5 \\
Std. deviation (adequacy)       & 1.42 \\
Std. deviation (fluency)        & 1.49 \\
Adequacy $\geq$ 3 (\%)          & 69.5\% \\
Fluency $\geq$ 3 (\%)           & 65.0\% \\
Adequacy--fluency correlation   & 0.865 \\
\hline
\end{tabular}
\caption{Overall human evaluation results (Model 5, 20 sentences, 11 evaluators)}
\label{tab:human_eval_overall}
\end{table}

The mean adequacy of 3.42 and fluency of 3.34 imply that, on average, evaluators rated the model’s translations above the midpoint of the scale, indicating at least partial but meaningful preservation of meaning and some degree of target-language naturalness. The high standard deviations (1.42 and 1.49) reflect a fair amount of variation across both sentences and evaluators; native speakers of Assyrian differ in their exposure to Classical Syriac, their familiarity with the Biblical register, and their personal standards of grammaticality \cite{artstein-poesio-2008-survey}. The very strong correlation between adequacy and fluency (r = 0.865) indicates that evaluators tended to rate the same translations as both adequate and fluent, or both inadequate and disfluent, suggesting the two dimensions are tightly coupled in this context.

At the sentence level, three sentences achieved an average overall score at or above 4.0 (``Moses was born in Egypt'', 4.27; ``Jesus rose from death three days after'', 4.09; ``Peace rested upon the valley'', 4.05), all notably short, syntactically simple sentences built from high-frequency Biblical vocabulary. The lowest-scoring sentence (``God promised Abraham that his descendants would be as many as the stars'', 2.68) involves a complex embedded clause and a metaphorical comparison that the phrase-based model struggles to handle, consistent with the known limitations of SMT on complex syntactic and semantic structures \cite{koehn2003statistical}.

\begin{figure}[htbp]
    \centering
    \includegraphics[width=0.98\textwidth]{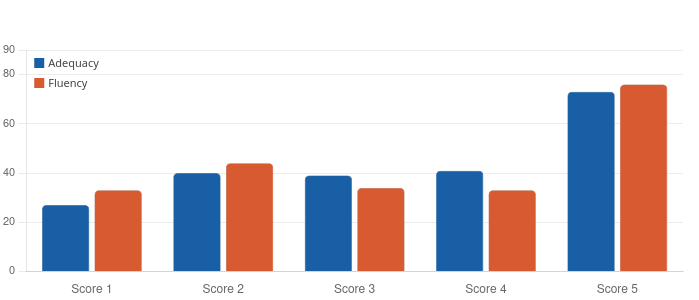}
    \caption{Distribution of adequacy and fluency ratings across all 220 judgments}
    \label{fig:dis_adc_flu}
\end{figure}

The distribution of individual ratings, shown in Figure~\ref{fig:dis_adc_flu}, reveals a bimodal pattern: score 5 is the single most frequent rating for both dimensions (33.2\% for adequacy, 34.5\% for fluency), alongside a notable tail of low scores. This bimodality is characteristic of phrase-based SMT output in low-resource settings, where translations of short, high-frequency phrases are often excellent while translations of longer or more complex sentences are poor \cite{kakum_22_nyishi}.

\subsection{Comparison with Related Work}
The best word-level BLEU achieved in this study is 23.54 (Model 5). Table~\ref{tab:comparison} places this result in context relative to comparable low-resource phrase-based SMT studies using Moses, Biblical or similarly domain-restricted corpora, and low-resource or morphologically rich target languages.

\begin{table}[h]
\centering
\small
\renewcommand{\arraystretch}{1.3}
\begin{tabular}{|l|l|c|c|}
\hline
\rowcolor{headergray}
\textbf{Study} & \textbf{Language Pair} & \textbf{Corpus Size} & \textbf{Best BLEU} \\
\hline
This study (Model 5)     & English $\rightarrow$ Syriac (Assyrian) & $\sim$36k pairs & \textbf{23.54} \\
\newcite{Chaya_2024_AFCM} & Aramaic $\rightarrow$ Hebrew            & $\sim$23k pairs & 44.30 \\
\newcite{kakum_22_nyishi}   & English $\rightarrow$ Nyishi            & 30k pairs       & 0.18 \\
\newcite{2016ialp.conf...20T} & English $\rightarrow$ Tigrinya          & $\sim$31k pairs & $\sim$20 \\
\hline
\end{tabular}
\caption{Comparison with related low-resource phrase-based SMT results}
\label{tab:comparison}
\end{table}

The score of 23.54 compares favorably with comparable low-resource SMT baselines on morphologically rich languages trained on Biblical corpora. \newcite{2016ialp.conf...20T}, working on English--Tigrinya, another Semitic language with a comparable Bible-derived corpus, report BLEU scores in the low-to-mid twenties using Moses with morphological segmentation, placing our result in a directly comparable range. The substantially higher 44.30 reported by \newcite{Chaya_2024_AFCM} for Aramaic--Hebrew reflects the linguistic proximity of that pair: shared script, largely shared vocabulary, and closely related morphology make alignment substantially easier than for the typologically distant English--Syriac pair. The very low BLEU reported for English--Nyishi underscores that BLEU scores are not comparable across language pairs, only within them \cite{papineni-etal-2002-bleu}.

\subsection{Analysis of Model Behavior}
Three patterns emerge across the six models. First, \textbf{data volume is the primary ceiling}: the 4.73-point drop from the 95/5 to the 70/30 split demonstrates that training data size is the dominant factor in translation quality, consistent with \newcite{velayuthan-etal-2024-back} and with the broader observation that phrase tables become sparser and less reliable as training data shrinks \cite{koehn-knowles-2017-six}. Second, \textbf{the OSM contributes meaningfully}: the +1.00 BLEU gain in Model 5 is the single most effective improvement in this study, providing the decoder with an explicit reordering signal for the English--Syriac word-order divergence \cite{durrani-etal-2011-joint}. Third, \textbf{higher distortion and larger BPE vocabularies offer diminishing returns}: Model 6's negligible gain suggests that the corpus's single-domain coverage is the binding constraint rather than reordering capacity or vocabulary coverage. The practical ceiling for Moses SMT on this corpus is approximately 23.5 BLEU, and pushing beyond it will require either additional parallel data or a transition to neural MT with cross-lingual transfer from a related higher-resource language such as Arabic \cite{boujkian2025improvinglowresourcemachinetranslation}.

\subsection{Preliminary NMT Experiments}
As a preliminary investigation into neural alternatives, fine-tuning experiments were conducted using two large multilingual NMT models, mBART-50 and NLLB-200, both fine-tuned on the same 95/5 data partition used for Models 1 and 5, with training taking approximately 43 hours per model on cloud GPU infrastructure. Both models produced empty output at inference time and could not be evaluated. The root cause was identified as the absence of Syriac (\texttt{syr\_Syrc}) from both models' pre-trained vocabularies: without a valid target-language token to condition generation on, the decoder had no starting point and produced no output regardless of the input. These experiments are reported as a negative result; resolving the issue requires adding \texttt{syr\_Syrc} as a new token, initialising its embedding from Arabic, and setting \texttt{forced\_bos\_token\_id} correctly, as outlined in Section~\ref{sec:conclusion}. The result reinforces the finding, consistent across the low-resource literature \cite{koehn-knowles-2017-six,Chaya_2024_AFCM,Guellil_2007}, that SMT remains the practical choice for language pairs absent from pre-trained multilingual vocabularies.

\subsection{Limitations}
Several limitations of this study should be acknowledged. The corpus is restricted entirely to Biblical text, so the model has no exposure to modern vernacular Assyrian, secular vocabulary, or non-liturgical discourse structures; its scores reflect performance within a narrow domain and should not be interpreted as measures of general-purpose translation quality. The single-reference BLEU evaluation underestimates true translation quality, as correct translations that differ from the reference in surface form, a common occurrence in morphologically rich languages, receive no credit \cite{papineni-etal-2002-bleu}. The phrase-based architecture cannot capture deep semantic nuances or complex metaphors as effectively as neural architectures. Finally, the human evaluation panel, while sufficient for inter-annotator agreement estimation \cite{artstein-poesio-2008-survey}, evaluates only the output of the best model and does not directly compare quality across models.

\section{Conclusion and Future Work}
\label{sec:conclusion}

This research presents the first published SMT model for the English-to-Syriac language pair, built using the Moses phrase-based framework and trained on a parallel corpus of 38,847 sentence pairs extracted from the complete English and Syriac Bible. The study shows that a functioning and reasonably accurate MT system can be created for this endangered low-resource language using publicly available tools and a corpus assembled from scratch, and it establishes quantitative baselines against which future systems for this language pair can be measured.

Across six model configurations, the outcome showed a reasonable translation quality, with word-level BLEU declining from 22.54 at the 95/5 split to 17.81 at the 70/30 split. The best configuration combines a 5-gram KenLM language model with an Operation Sequence Model for explicit reordering, achieving a word-level BLEU of 23.54; the +1.00 improvement over the identical-split baseline reflects the genuine benefit of addressing the SVO-to-VS structural divergence between English and Syriac. Human evaluation by eleven native Assyrian speakers produced mean adequacy and fluency scores of 3.42 and 3.34 out of 5, with strong performance on short, high-frequency constructions and weak performance on complex ones, the defining characteristic of phrase-based SMT in low-resource settings. Beyond the translation model, this work contributes a verse-aligned, manually verified parallel corpus prepared for public release, and a collection of secular Syriac documents from the Ministry of Education being processed for release through the same pipeline and available privately to researchers who request them for academic use.

Future work could be summarized as follows. First, expanding the parallel corpus beyond Biblical texts: the Ministry of Education documents will introduce secular vocabulary and register diversity, and sources such as the Tatoeba English--Assyrian dataset and the NENA (North-Eastern Neo-Aramaic) Corpus represent near-term options. Second, replacing language-agnostic BPE with a morphology-aware segmenter tailored to Syriac's root-and-pattern system, following the significant BLEU gains shown for the closely comparable English--Tigrinya pair \cite{2016ialp.conf...20T}. Third, completing the NMT comparison: adding \texttt{syr\_Syrc} to the vocabularies of NLLB-200 and mBART-50, initializing its embedding from Arabic, would enable a direct neural comparison against the SMT baselines established here, and in the longer run, cross-lingual transfer from Arabic represents the most promising path toward a general-purpose system \cite{boujkian2025improvinglowresourcemachinetranslation}. Fourth, evaluation should move to multi-reference scoring to better handle surface-form variation in Syriac, and the human evaluation panel could be expanded to include speakers with different levels of familiarity with the language.

\section*{Ethical Consideration}
The collected Biblical data are publicly available. The data from the Ministry of Education (the Kurdistan Regional Government in the Kurdistan Region of Iraq) could be available upon request. 

\section*{Author Contributions}
Conceptualization, Hadiana Sliwa (H.S.) and Hossein Hassani (H.H.); methodology, H.S. and H.H.; software, H.S.; validation, H.S.; formal analysis, H.S.; investigation, H.S.; resources, H.S. and H.H.; data curation, H.S.; preparing first draft, H.S.; revising the draft and preparing final manuscript, H.H.; visualization, H.S.; supervision, H.H.; project administration, H.H.

\section*{Dataset Availability}
The dataset is available \href{https://github.com/hadianasliwa/MT_between_English-Syriac_using_SMT}{here}. 

\section*{Acknowledgments}
We thank the two collaborators who participated in the manual alignment review, the eleven native Assyrian speakers who participated in the human evaluation, and the Assyrian-related department of the Ministry of Education (the Kurdistan Regional Government in the Kurdistan Region of Iraq) for providing secular Syriac documents and supporting their use in this research.

\bibliographystyle{lrec}
\bibliography{references}

\end{document}